\documentclass{article}

\usepackage{arxiv}

\usepackage[utf8]{inputenc} 
\usepackage[T1]{fontenc}    
\usepackage{hyperref}       
\usepackage{url}            
\usepackage{booktabs}       
\usepackage{amsfonts}       
\usepackage{nicefrac}       
\usepackage{microtype}      
\usepackage{graphicx}
\usepackage[numbers]{natbib}
\usepackage{doi}
\usepackage{float}
\usepackage{placeins}
\usepackage{needspace}
\usepackage{amsmath}
\usepackage{listings}
\usepackage{xcolor}
\usepackage{subcaption}
\usepackage{tcolorbox}

\usepackage{fancyhdr}
\tcbuselibrary{listings,skins,breakable}

\newtcblisting{promptbox}[1][]{
  colback=gray!5!white,
  colframe=gray!75!black,
  listing only,
  listing options={basicstyle=\ttfamily\small,breaklines=true},
  title={#1},
  breakable,
  enhanced,
  sharp corners,
  boxrule=0.5pt
}

\title{Alvorada-Bench: Can Language Models Solve Brazilian University Entrance Exams?}

\author{
  Henrique Godoy \\
  Inteli \\
  São Paulo, Brazil \\
  henrique.godoy@sou.inteli.edu.br
}

\date{}

\hypersetup{
pdftitle={Alvorada-Bench: Can Language Models Solve Brazilian University Entrance Exams?},
pdfsubject={cs.CL, cs.AI},
pdfauthor={Henrique Godoy}
}

\begin{document}
\maketitle

\begin{abstract}
Language models are increasingly used in Brazil, but most evaluation remains English-centric. This paper presents Alvorada-Bench \footnote{Data and code available at \url{https://huggingface.co/datasets/HenriqueGodoy/Alvorada-bench} and \url{https://github.com/herniqeu/Alvorada-bench}}, a 4,515-question, text-only benchmark drawn from five Brazilian university entrance examinations. Evaluating twenty models under zero-shot, role-playing, and chain-of-thought prompting, producing 270,900 responses with structured self-reports of confidence, perceived difficulty, and Bloom level. The top models exceed 94\% accuracy overall, but accuracy declines on Mathematics and on the engineering oriented IME and ITA exams, indicating persistent weaknesses in multi-step reasoning. Confidence is well calibrated and correlates with perceived difficulty, revealing that models can accurately assess their own certainty capabilities. A cost-accuracy analysis shows that high accuracy is achievable at under \$2 per 1K tokens. On ENEM 2024 the top model (O3) achieved perfect scores in Languages subject questions while even the weakest system (GPT-4.1 Nano) only underperforms humans in Mathematics. Through exams that distill decades of Brazilian educational priorities and assess millions of students yearly, Alvorada-Bench establishes whether language models can navigate the intersection of language, culture, and reasoning that defines academic readiness in Brazil.
\end{abstract}

\section{Introduction}

Language models increasingly mediate critical decisions across diverse applications, from educational assessment to medical diagnosis, yet their evaluation remains predominantly English-centric. As these models expand into global markets serving linguistically and culturally diverse populations, this evaluation gap poses significant risks. 

Current evaluations demonstrate remarkable performance on standardized tests. GPT-4 scores at the 90th percentile on the SAT, passes the Bar Exam in the top 10\%, and outperforms 85\% of participants in coding contests \cite{openai2024gpt4technicalreport,hou2024comparing}. However, these benchmarks embed cultural assumptions that limit global applicability. Translation cannot address implicit cultural frameworks, as SAT questions about financial aid assume familiarity with concepts irrelevant in countries with free universities. This cultural specificity produces measurable degradation: performance drops from 70.9\% on English educational tasks to 49.7\% in Telugu \cite{gupta2025multilingualperformancebiaseslarge}, while Chinese outputs exhibit 41\% lexical divergence from native usage despite English-like syntactic patterns \cite{guo2025largelanguagemodelsenglish}.

These issues are evident in non-English contexts such as Brazil, where Portuguese serves a population exceeding 220 million, making it the sixth most spoken language worldwide. However, Portuguese remains underrepresented in the benchmarks. Brazilian university entrance exams offer a compelling solution that combines cultural specificity with rigorous standardization. Refined over decades through expert review, statistical validation, and millions of student responses, these exams serve as a natural experiment in knowledge assessment, capturing both cognitive demands and the cultural knowledge expected of educated Brazilians.

To address this gap, this paper introduce Alvorada-Bench: a benchmark comprising 4,515 questions drawn from five Brazilian university entrance examinations—ENEM (Exame Nacional do Ensino Médio), FUVEST (São Paulo), UNICAMP (Campinas), IME (Instituto Militar de Engenharia), and ITA (Instituto Tecnológico de Aeronáutica) spanning from 1981 to 2025. Using Alvorada-Bench, we conduct a controlled evaluation of 20 models from OpenAI, Anthropic, and DeepSeek under zero-shot, role-playing, and chain-of-thought prompting strategies.

Recent work \cite{almeida2023bluexbenchmarkbasedbrazilian} introduced BLUEX, a 1,095-question corpus drawn from UNICAMP and USP exams, providing early evidence that Brazilian exams can serve as evaluation substrates. Another study \cite{locatelli2024examiningbehaviorllmarchitectures} examined language model behavior on Brazilian standardized exams and provided human performance baseline data. Although Alvorada-Bench was constructed independently and did not reuse BLUEX items, these studies motivate and contextualize this work.

This work contributes three elements: (1) Alvorada-Bench, a benchmark of 4,515 questions compiled from five Brazilian entrance examinations (ENEM, FUVEST, UNICAMP, IME, ITA) spanning 1981 to 2025 and covering four disciplinary areas aligned with the BNCC; (2) a controlled evaluation of 20 language models that yields 270,900 model question interactions; and (3) an empirical analysis of calibration, cost efficiency, and cognitive complexity profiles, encompassing model uncertainty quantification, subject and exam level performance patterns, prompt strategy effects, and stratification by Bloom's taxonomy.

\section{Dataset and Methodology}

\subsection{The Alvorada-Bench Dataset}

Alvorada-Bench comprises 4,515 multiple-choice questions extracted from Brazilian university entrance examinations, collected from 126 test administrations spanning 1981-2025. The dataset integrates questions from five distinct examination systems that collectively assess over 5 million Brazilian students annually. Table 1 presents the distribution across examination sources: ENEM contributes 1,629 questions (36.1\%), FUVEST 1,303 (28.9\%), UNICAMP 716 (15.9\%), ITA 720 (15.9\%), and IME 147 (3.3\%). This distribution reflects the relative importance of each examination within the Brazilian higher education admission system, where ENEM functions as the national standardized assessment while FUVEST, UNICAMP, ITA, and IME serve as selective admissions instruments for specific institutions.

\begin{table}[htbp]
\centering
\begin{tabular}{lllll}
\toprule
Examination & Questions & Percentage & Years Covered & Sessions\\
\midrule
ENEM & 1,629 & 36.1\% & 2010-2024 & 26 \\
FUVEST & 1,303 & 28.9\% & 1981-2025 & 32 \\
IME & 147 & 3.3\% & 2017-2023 & 7 \\
ITA & 720 & 15.9\% & 2008-2024 & 46 \\
UNICAMP & 716 & 15.9\% & 2011-2025 & 15 \\
\textbf{Total} & \textbf{4,515} & \textbf{100.0\%} & \textbf{1981-2025} & \textbf{126} \\
\bottomrule
\end{tabular}
\caption{Dataset Composition by Examination Source}
\label{tab:dataset_composition}
\end{table}

The dataset spans four major disciplinary categories aligned with the Brazilian National Curriculum Base (Base Nacional Comum Curricular - BNCC). Natural Sciences constitutes 36.9\% of the dataset (1,667 questions), encompassing Chemistry, Physics, and Biology. Human Sciences represents 28.2\% (1,275 questions), covering History, Geography, Sociology, and Philosophy. Languages comprises 18.0\% (814 questions), with Portuguese Language as the primary component supplemented by English and Spanish assessments. Mathematics accounts for 16.8\% (759 questions), testing quantitative reasoning and problem-solving capabilities.

\begin{figure}[htbp]
\centering
\includegraphics[width=0.85\textwidth]{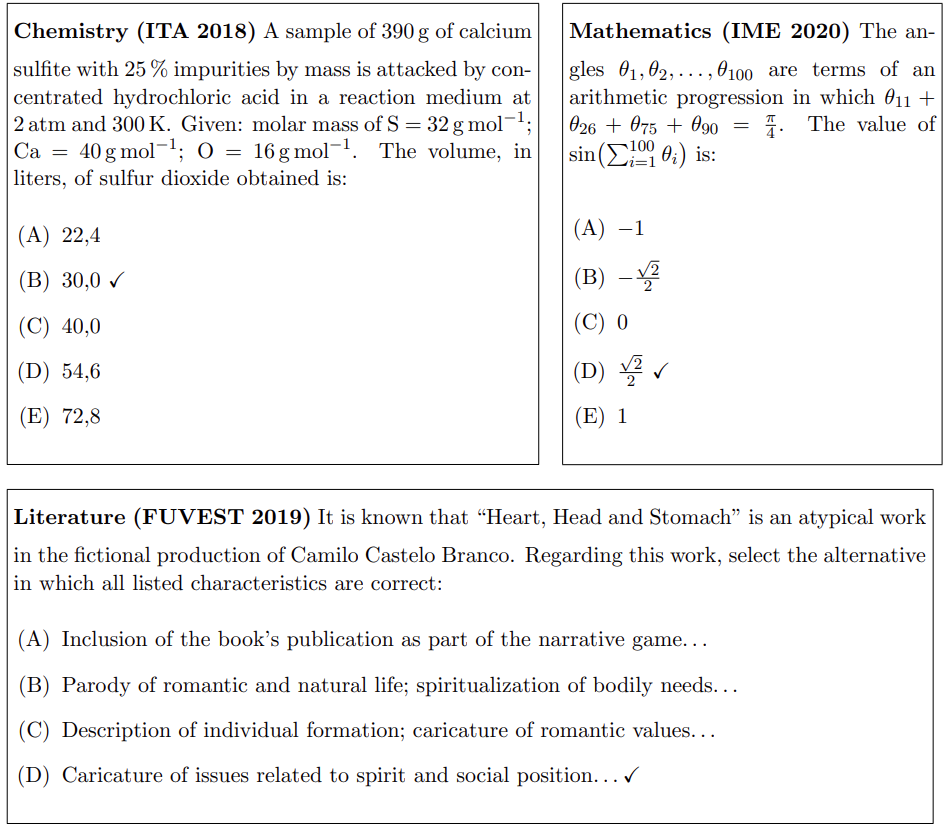}
\caption{Representative Question Examples from Alvorada-Bench}
\label{fig:questions}
\end{figure}

\FloatBarrier

Figure~\ref{fig:questions} illustrates the diversity of questions across different examinations and subject areas. The Chemistry question (ITA 2018) requires stoichiometric calculations involving calcium sulfate reactions with hydrochloric acid, demonstrating the quantitative reasoning expected in engineering entrance exams. The Mathematics question (IME 2020) combines geometric concepts with arithmetic progressions, requiring multi-step algebraic manipulation. The Literature question (FUVEST 2019) tests comprehension of 19th-century Portuguese literature, specifically Camilo Castelo Branco's "Heart, Head and Stomach," demanding familiarity with romantic literary conventions and the ability to identify thematic elements within Brazilian-Portuguese cultural context.

\subsection{Dataset Construction}

The dataset was built through a systematic pipeline designed to preserve question integrity while ensuring compatibility with text-based model evaluation. All examination materials were acquired in PDF format with their corresponding official answer keys, providing authoritative ground truth for evaluation.

\begin{figure}[htbp]
\centering
\includegraphics[width=1.0\textwidth]{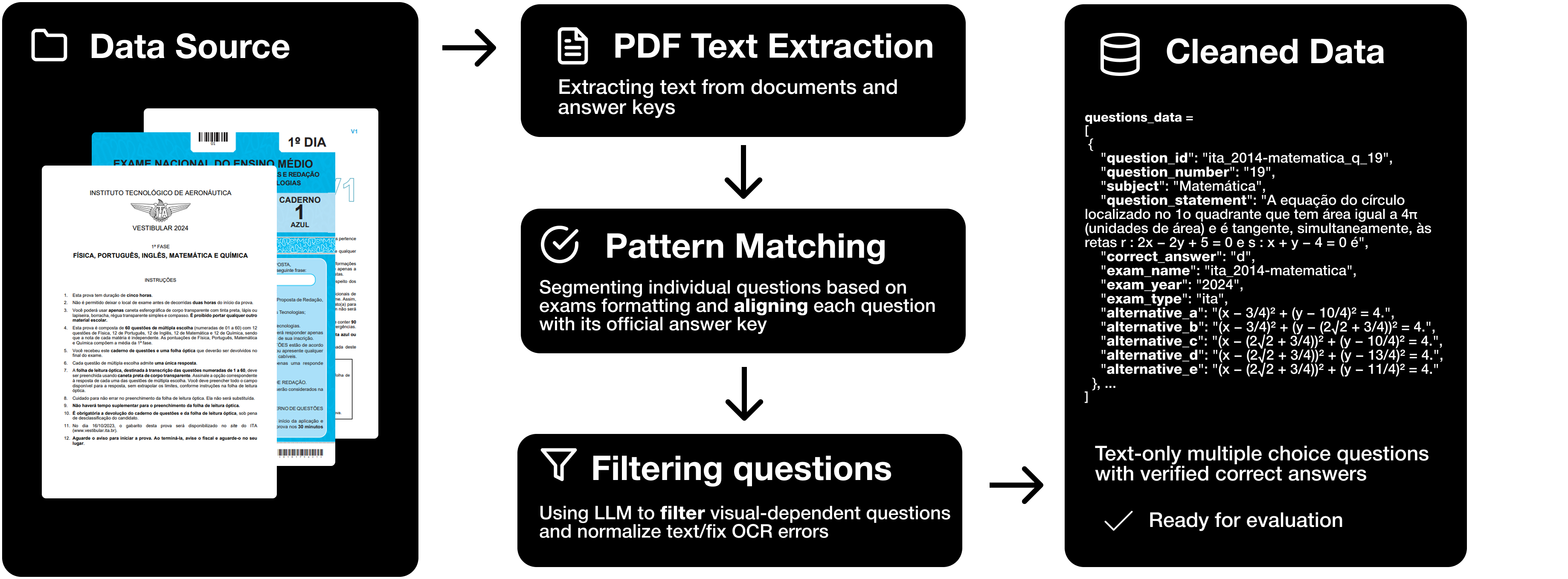}
\caption{Dataset Construction Pipeline}
\label{fig:dataset_pipeline}
\end{figure}

\FloatBarrier

The construction pipeline, illustrated in Figure~\ref{fig:dataset_pipeline}, consists of four sequential processing stages. First, PDF text extraction processes the examination documents from various years and sources. Second, pattern matching employs regular expression techniques to identify question boundaries by detecting structural regularities in Brazilian examination formatting: sequential numbering patterns, multiple-choice alternatives (typically five options labeled A through E, with the exception of UNICAMP which uses four options A through D), and consistent typographical markers that delineate question start and end points. Each segmented question undergoes automated alignment with official answer keys to establish the correct responses.

Third, the filtering stage processes the questions in batches through a language model to identify and exclude items incompatible with text-only evaluation. The language model analyzes each batch to detect questions requiring visual interpretation (graphs, diagrams, maps, geometric figures). Finally, text normalization addresses formatting inconsistencies identified during the filtering process and ensures the proper preservation of mathematical notation and chemical formulae. The pipeline output consists of 4,515 validated text-only multiple-choice questions with verified correct answers, ready for language model evaluation.

\subsection{Evaluation Methodology}

This work evaluated twenty language models representing diverse architectures and training approaches. The models were accessed through official APIs in Aug 2025 from three major providers: OpenAI (twelve models), Anthropic (six models) and DeepSeek (two models).

Each prompting strategy required structured output in JSON format containing four elements: selected answer (constrained to alternatives A-E), confidence score (integer scale 0–10), perceived difficulty rating (integer scale 0–10), and Bloom's taxonomy classification (remember, understand, apply, analyze, evaluate or create). This structured output enabled quantitative analysis of both performance metrics and metacognitive assessments, facilitating investigation of calibration quality and systematic biases in difficulty perception.

\section{Results}

Results are presented hierarchically, beginning with overall accuracy, cost-efficiency and progressing to calibration, subject-level analyses, exam-type variation, prompting effects, and cognitive complexity profiling.

\subsection{Model Performance Overview}

\begin{table}[htbp]
\centering
\begin{tabular}{lll}
\toprule
Model & Accuracy & Relative to Mean \\
\midrule
O3 Pro & 0.9463 & 0.133075 \\
O3 & 0.9455 & 0.132275 \\
O1 & 0.9308 & 0.117575 \\
DeepSeek Reasoner & 0.9271 & 0.113875 \\
O4 Mini & 0.9150 & 0.101775 \\
O1 Preview & 0.9148 & 0.101575 \\
O3 Mini & 0.8815 & 0.068275 \\
Claude Opus 4 & 0.8674 & 0.054175 \\
Claude Sonnet 4 & 0.8346 & 0.021375 \\
O1 Mini & 0.8203 & 0.007075 \\
Claude 3.7 Sonnet & 0.7990 & -0.014225 \\
Claude 3.5 Sonnet & 0.7941 & -0.019125 \\
DeepSeek Chat & 0.7912 & -0.022025 \\
Claude 3 Opus & 0.7644 & -0.048825 \\
GPT-4.1 & 0.7499 & -0.063325 \\
GPT-4o & 0.7363 & -0.076925 \\
GPT-4.1 Mini & 0.7155 & -0.097725 \\
Claude 3.5 Haiku & 0.6763 & -0.136925 \\
GPT-4o Mini & 0.6496 & -0.163625 \\
GPT-4.1 Nano & 0.6049 & -0.208325 \\
\bottomrule
\end{tabular}
\caption{Model Performance Rankings}
\label{tab:model_performance}
\end{table}

\FloatBarrier

Table~\ref{tab:model_performance} ranks the evaluated systems by mean accuracy. Reasoning-optimised models dominate: \textbf{O3 Pro (94.63\%)}, \textbf{O3 (94.55\%)}, and \textbf{O1 (93.08\%)} lead the leaderboard, outperforming the corpus mean (81.33\%) by 11–13 percentage points. DeepSeek Reasoner (92.71\%) and O4 Mini (91.50\%) round out the top five. The accuracy gap between the best (O3 Pro) and worst (GPT-4.1 Nano, 60.49\%) systems spans \textbf{34.1 percentage points (p.p.)}, illustrating pronounced stratification within current LLM offerings. 

\FloatBarrier

\subsection{Comparison with Human Performance Baselines}

\begin{figure}[htbp]
\centering
\includegraphics[width=0.7\textwidth]{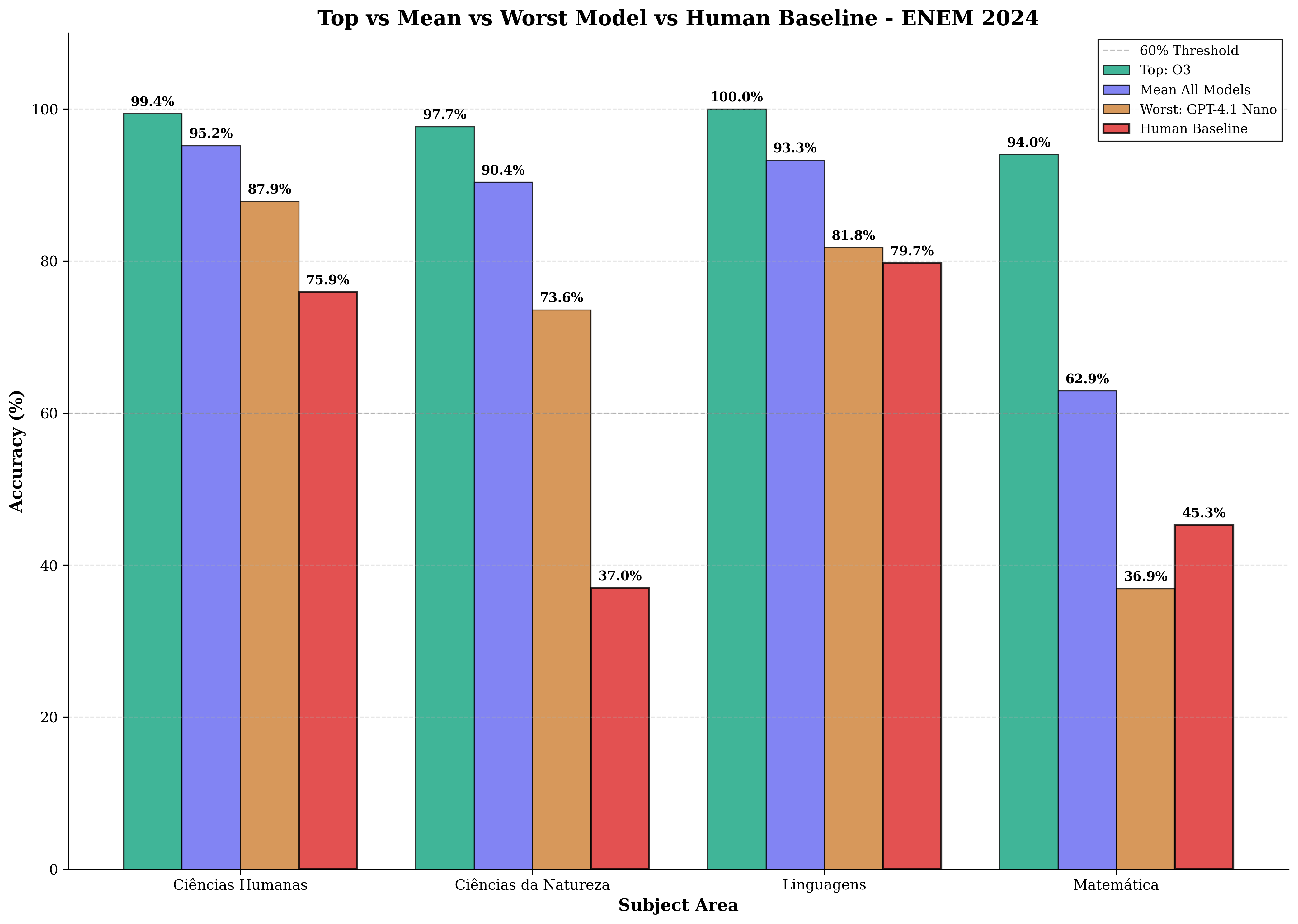}
\caption{Comparison of LLM models performance with Brazilian student baseline on ENEM 2024 across different domains.}
\label{fig:human_baseline}
\end{figure}

Figure~3 juxtaposes language model performance with Brazilian student outcomes on the \textbf{ENEM 2024} exam, which evaluated more than 4 million Brazilian students seeking university admission. The human baseline data is derived from \cite{locatelli2024examiningbehaviorllmarchitectures}. Our results demonstrate that language models now systematically outperform Brazilian students in most ENEM 2024 domains, with the top model (O3) achieving perfect scores in Languages while even the weakest system (GPT-4.1 Nano) only underperforms humans in Mathematics, marking a decisive shift in the human--language model capability balance on standardized educational assessments. All 20 models surpass the human baseline in Humanities, Natural Sciences, and Languages. The top model (O3) achieves a flawless 100\% on Languages and maintains $\geq 94\%$ across domains, highlighting the rapid closing of human--language model performance gaps on curriculum-level tasks.

\FloatBarrier
\subsection{Cost-Efficiency Frontier and Temporal Evolution}

\begin{figure}[htbp]
\centering
\includegraphics[width=1.0\textwidth]{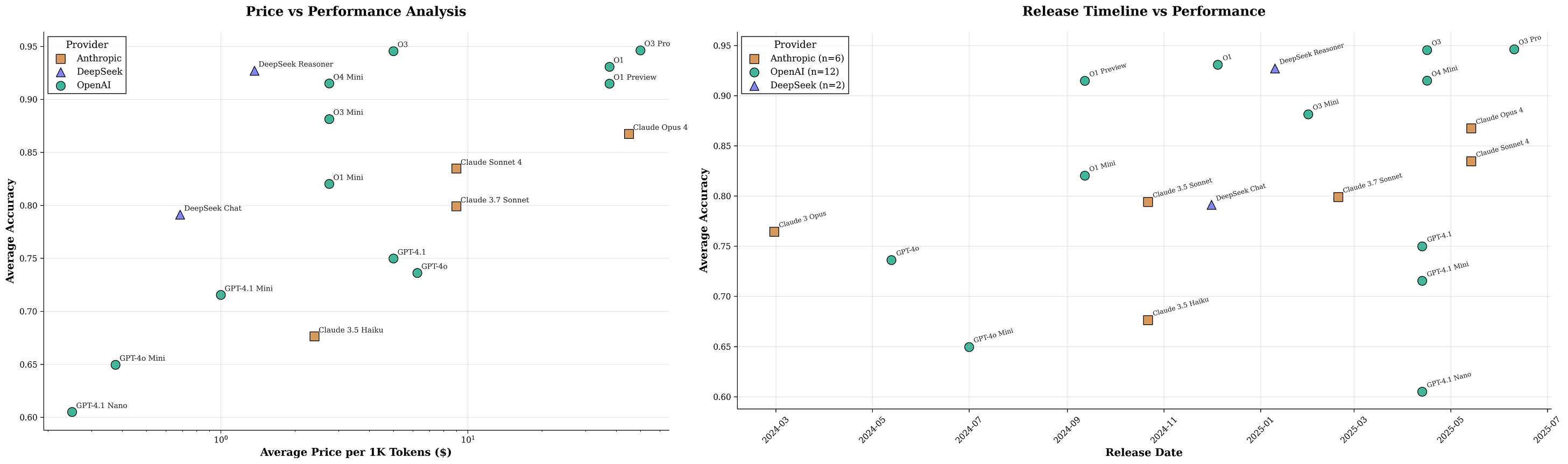}
\caption{(a) Cost-efficiency frontier: price per 1K tokens vs accuracy. (b) Temporal evolution of model performance (2024-2025).}
\label{fig:price_perfomance_date}
\end{figure}

Figure~\ref{fig:price_perfomance_date} shows the cost-efficiency frontier and temporal evolution. The cost-efficiency frontier reveals that high performance no longer requires premium pricing. DeepSeek Reasoner and O3 Mini deliver more accuracy 91\% at less than \$2 per 1K tokens, while expensive models such as GPT-4.1 (\$15) offer diminishing returns, democratizing access to near-state-of-the-art capabilities. \textbf{DeepSeek Reasoner (92.71\%, \$1.82)} and \textbf{O3 Mini (91.50\%, \$1.95)} dominate the cost–accuracy frontier, offering near-state-of-the-art precision at $<$ \$2. In contrast, GPT-4.1 (\$15.00) trails the frontier by 3 percentage points, underscoring diminishing returns at higher price points.

Temporal analysis reveals a dramatic acceleration in model capabilities that occurred in Q2 2024, coinciding with the public release of reasoning-supervised architectures. Leading accuracy climbed from 73.6\% (GPT-4o, May 2024) to 94.6\% (O3 Pro, the leading system).

\FloatBarrier
\subsection{Model Calibration and Uncertainty Quantification}

\FloatBarrier
\begin{figure}[htbp]
\centering
\includegraphics[width=1.0\textwidth]{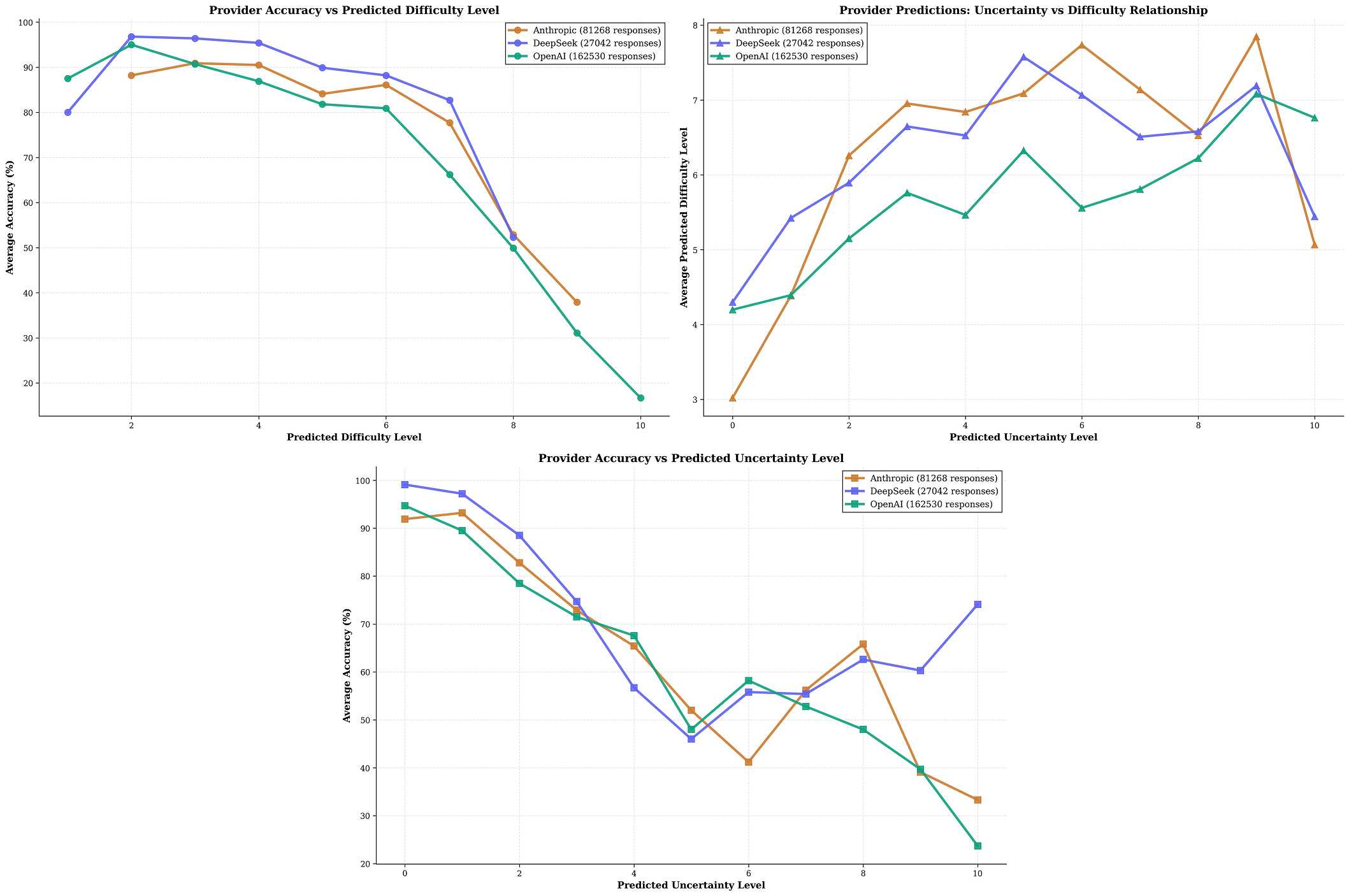}
\caption{(a) Model calibration: predicted vs actual accuracy. (b) Accuracy vs self-reported uncertainty. (c) Uncertainty vs perceived difficulty cor-
relation.}
\label{fig:uncertainty_difficulty_accuracy}
\end{figure}

\FloatBarrier

Beyond point accuracy, deployment scenarios require well-calibrated confidence estimates. Figure~\ref{fig:uncertainty_difficulty_accuracy} shows that modern LLMs demonstrate well-calibrated confidence: models accurately predict their own performance levels, with self-assessed high-confidence responses achieving $>$90\% accuracy, enabling reliable deployment in risk-sensitive applications. Self-reported uncertainty serves as a reliable performance indicator: models expressing high confidence consistently achieve higher accuracy, while uncertain responses flag potential errors, providing a practical mechanism for human oversight prioritization. Furthermore, models exhibit human-like metacognitive awareness: their uncertainty correlates positively with perceived question difficulty, suggesting that current LLMs can identify challenging problems and appropriately modulate confidence—a crucial capability for educational deployment. The bin-wise calibration curves reveal that responses labelled with low uncertainty (levels 0–1) exceed 90\% accuracy, and accuracy degrades monotonically with rising uncertainty. Complementary analysis demonstrates a positive correlation between self-reported uncertainty and predicted item difficulty, indicating that contemporary LLMs emit informative confidence signals suitable for risk-aware applications.

\FloatBarrier
\subsection{Subject-Level Performance Analysis}

\begin{figure}[htbp]
\centering
\begin{minipage}[t]{0.55\textwidth}
\vspace{0pt}
\centering
\includegraphics[width=\textwidth]{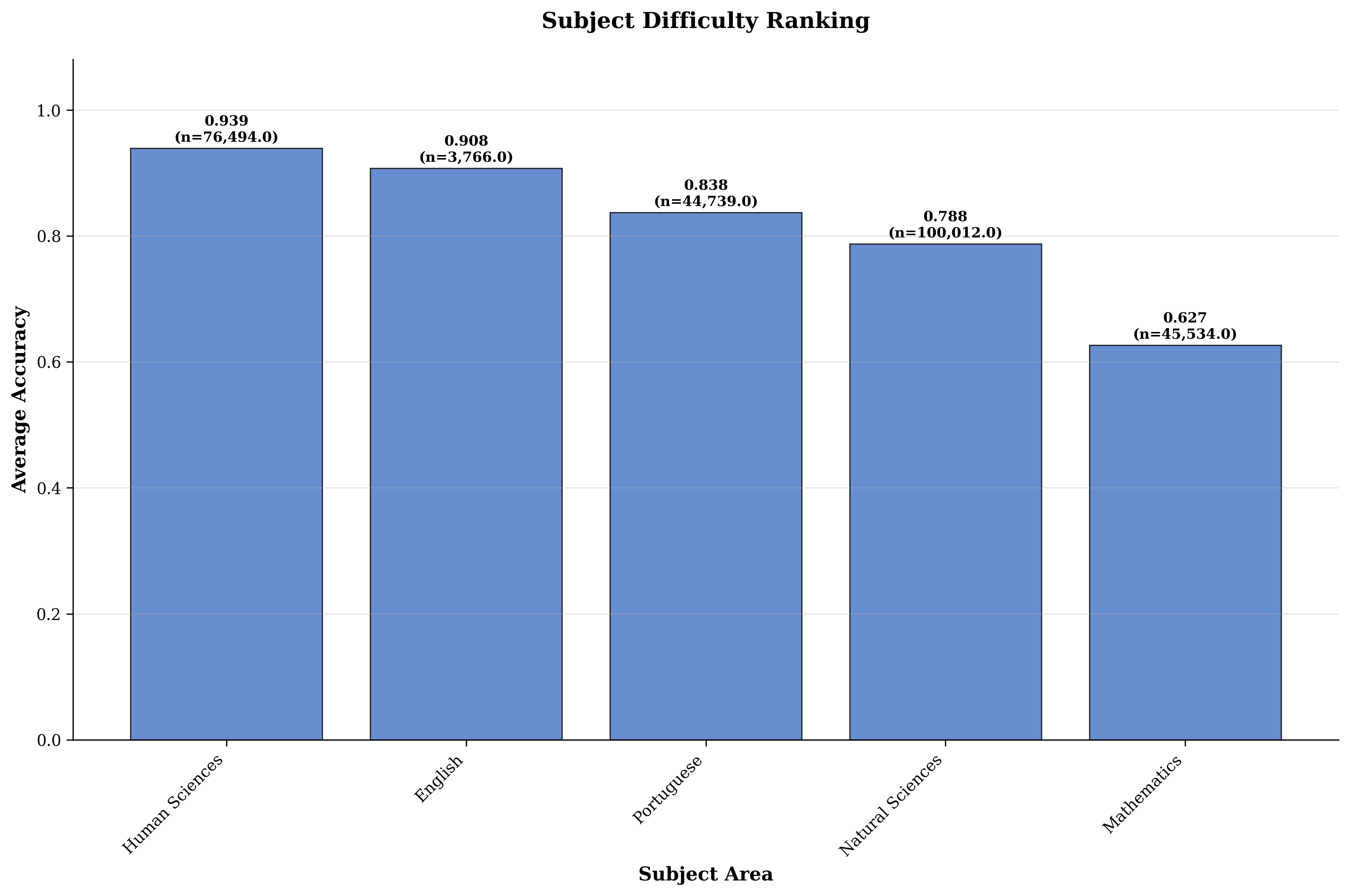}
(a) Subject difficulty ranking by model accuracy.

\vspace{0.5cm}
\small
\raggedright
Performance exhibits marked disparities across academic domains (Fig.~\ref{fig:subject_analysis}). Humanities disciplines achieve superior accuracy (Human Sciences 93.9\%, English 90.8\%), while quantitative fields significantly underperform (Mathematics 62.7\%). Reasoning-enhanced models substantially mitigate these deficiencies, with mathematics performance reaching 93.8\% for O3 and 93.7\% for DeepSeek Reasoner—representing improvements exceeding 48 percentage points relative to baseline models. Natural Sciences similarly benefits, with top reasoning models achieving 94.5\% (O3) compared to 70.4\% for standard Claude 3 Opus. Despite these advances, mathematical computation and symbolic reasoning persist as primary performance bottlenecks for conventional architectures.\end{minipage}
\hfill
\begin{minipage}[t]{0.4\textwidth}
\vspace{0pt}
\centering
\includegraphics[width=\textwidth]{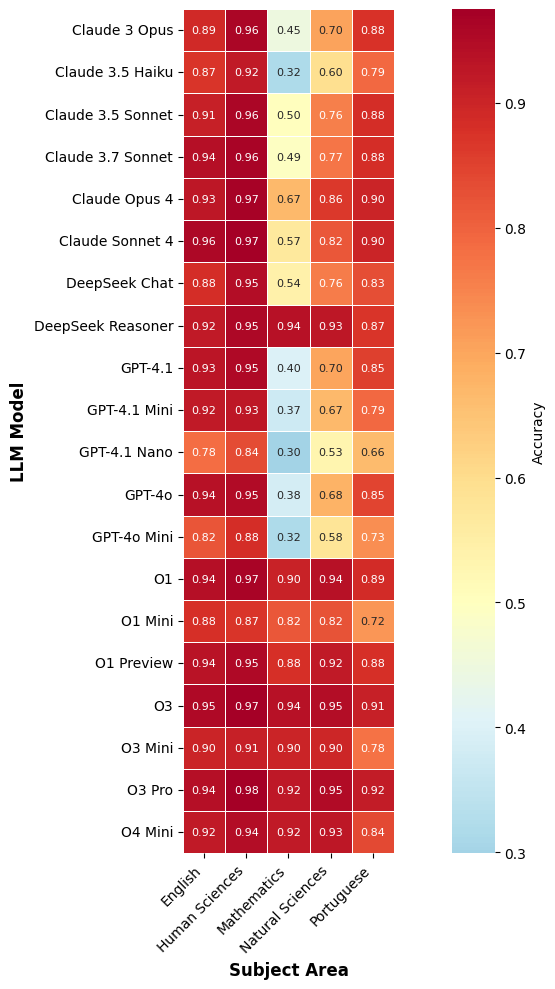}
(b) Performance comparison across subject areas.
\end{minipage}
\caption{Subject-level performance analysis across academic domains.}
\label{fig:subject_analysis}
\end{figure}

\FloatBarrier

\subsection{Examination-Type Performance Patterns}

\begin{figure}[htbp]
\centering
\includegraphics[width=0.65\textwidth]{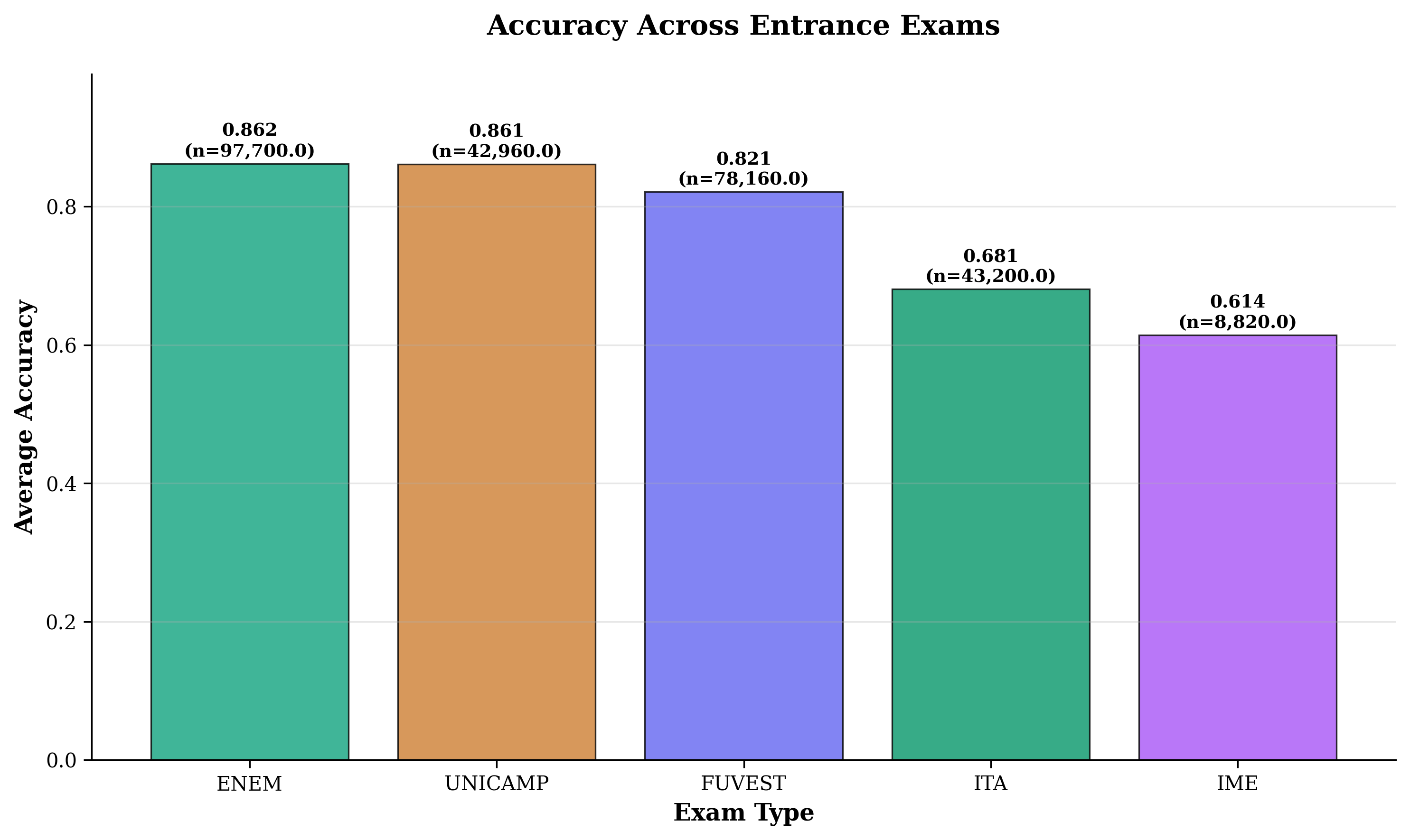}
\caption{Model performance across different Brazilian entrance exams.}
\label{fig:exam_performance}
\end{figure}

\FloatBarrier

Figure~7 reveals stratified performance patterns across the five entrance examinations. Models achieve comparable accuracy on comprehensive assessments (ENEM 86.2\%, UNICAMP 86.1\%), with modest degradation on FUVEST (82.1\%). Performance deteriorates substantially on specialized engineering examinations, dropping to 68.1\% for ITA and 61.4\% for IME, representing a 24.8 percentage point decline from ENEM. This stratification underscores a fundamental limitation: while current LLMs demonstrate robust performance on interdisciplinary evaluations, they exhibit marked deficiencies when confronting computation-intensive, domain-specific problem-solving tasks characteristic of technical entrance examinations.\Needspace{40\baselineskip}
\subsection{Prompt Engineering Effects}
\begin{figure}[H]
\centering
\begin{minipage}[t]{0.55\textwidth}
\vspace{0pt}
\centering
\includegraphics[width=\textwidth]{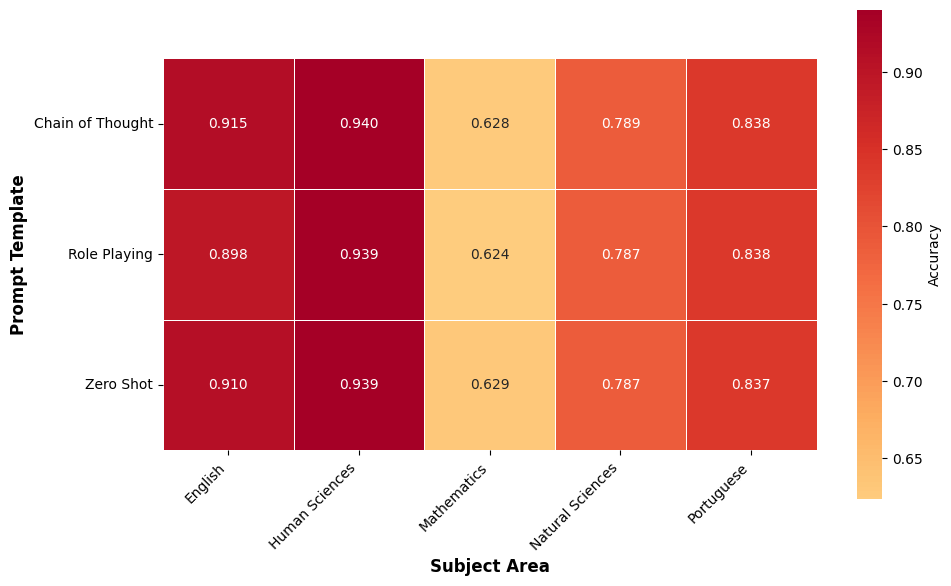}
(a) Prompt strategy performance across subject areas.

\vspace{0.3cm}
\small
\raggedright
This work evaluated three prompting paradigms: zero-shot, role-playing, and chain-of-thought. Across 20 models, accuracy variance remains minimal—typically under 1 percentage point within each model, with a maximum spread of 1.6 percentage points (Claude 3 Opus). Reasoning-optimized architectures demonstrate exceptional prompt invariance, with O3 exhibiting merely 0.1 percentage point variation and O1 showing 0.3 percentage point difference across paradigms. This consistency suggests that advanced reasoning capabilities confer inherent robustness to instruction framing, rendering prompt engineering largely redundant for these architectures.\end{minipage}
\hfill
\begin{minipage}[t]{0.43\textwidth}
\vspace{0pt}
\centering
\includegraphics[width=\textwidth]{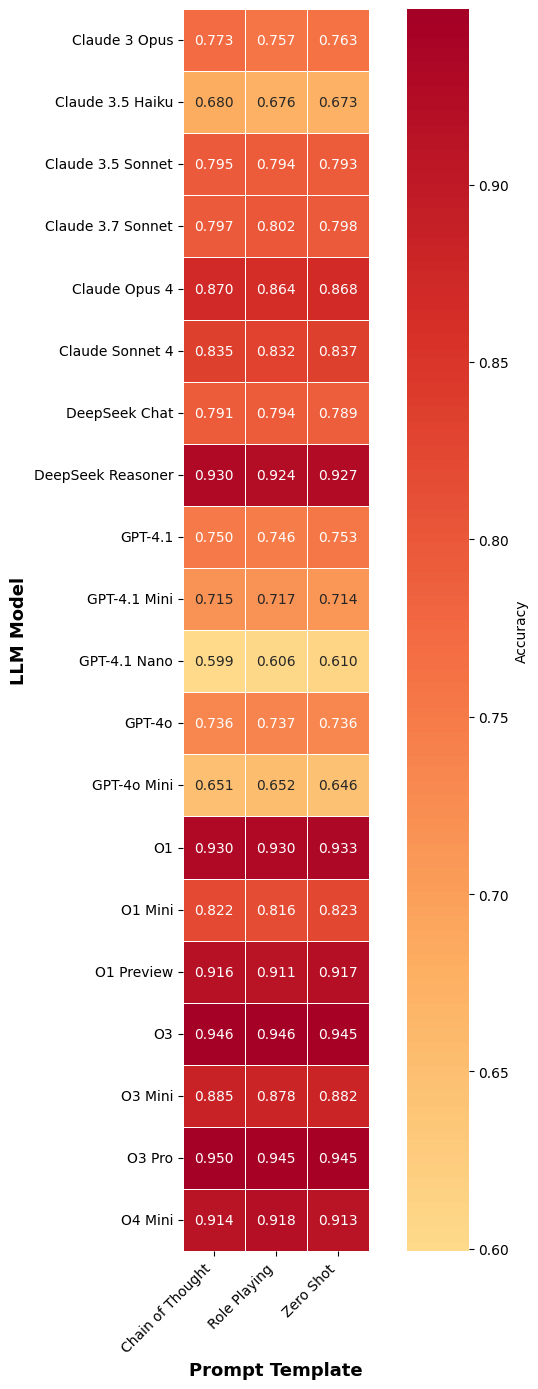}
(b) Model sensitivity to prompting strategies.
\end{minipage}
\caption{Analysis of prompt engineering effects across models and subjects.}
\label{fig:prompt_analysis}
\end{figure}

\FloatBarrier

\subsection{Cognitive-Complexity Profile}

\begin{figure}[htbp]
\centering
\begin{minipage}[t]{0.55\textwidth}
\vspace{0pt}
\small
\raggedright
Figure 9 stratifies performance across Bloom's cognitive taxonomy levels. Models demonstrate strong competence at knowledge retrieval (Remember: 92.4\% mean) and comprehension (Understand: 92.0\% mean), with surprisingly robust performance on evaluation tasks (87.8\% mean). However, application-level tasks emerge as the critical bottleneck, showing both the lowest mean accuracy (69.7\%) and highest variance across models—ranging from 39.8\% (GPT-4.1 Nano) to 97.6\% (O3 Pro). This non-monotonic pattern challenges conventional assumptions about cognitive hierarchies in LLMs. While reasoning-enhanced architectures (O3, O1, DeepSeek Reasoner) achieve near-parity across all taxonomic levels (>90\%), standard models exhibit a pronounced performance valley at the application tier, suggesting that translating conceptual understanding into practical problem-solving constitutes the primary challenge for current language models.
\end{minipage}
\hfill
\begin{minipage}[t]{0.43\textwidth}
\vspace{0pt}
\centering
\includegraphics[width=\textwidth]{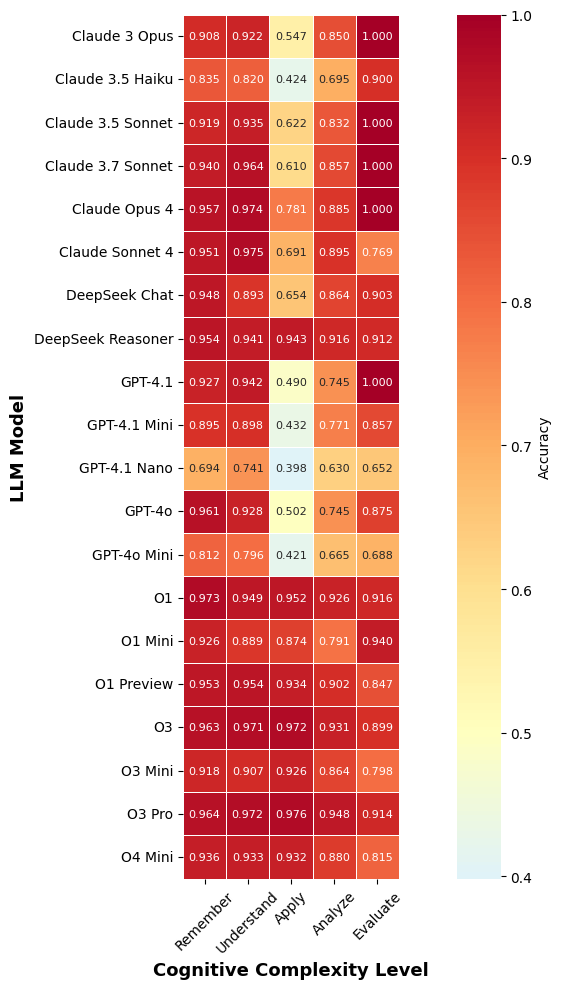}
\end{minipage}
\caption{Model accuracy stratified by Bloom's taxonomy.}
\label{fig:bloom}
\end{figure}

\FloatBarrier

\section{Discussion}

Brazilian university entrance examinations expose language models to cultural references and quantitative reasoning that are rarely co-located in existing benchmarks. The results show that systems now exceed 90 \% accuracy on humanities items rich in Brazilian historical and literary content, indicating substantial assimilation of culturally specific knowledge during pre-training. In contrast, accuracy falls below 70 \% on Mathematics, Physics, and Chemistry questions and declines further on the computation-heavy ITA and IME exams, confirming that symbolic manipulation and multi-step reasoning remain key failure modes. Confidence scores track empirical accuracy monotonically and correlate with perceived difficulty, suggesting that current models can signal when human oversight is warranted. At the same time, a shift in the cost–accuracy frontier models priced under \$2 per 1 K tokens achieve $\geq$ 92 \% accuracy lowers the barrier to large-scale educational deployment while raising questions about equitable access. Temporal analysis reveals a 21-point accuracy gain within one year, coinciding with the introduction of reasoning-supervised architectures, underscoring the role of targeted alignment over raw scale. Together, these patterns echo multilingual studies that report strong cultural knowledge yet persistent quantitative weaknesses.

\subsection{Limitations}

Several limitations and threats to validity warrant caution. First, our evaluation excludes multimodal questions requiring figures, diagrams, or maps, limiting assessment to 4,515 text-only multiple-choice items. The evaluation scores only final answers without evaluating intermediate reasoning steps, potentially overestimating performance when models arrive at correct answers through spurious correlations. This binary grading differs from human evaluation where partial credit rewards correct methodology.

Data contamination presents an inherent risk since Brazilian entrance exams are publicly available. Without provider decontamination reports, our accuracy measurements represent upper bounds on true out-of-distribution performance. The structured output format requiring confidence scores and difficulty assessments may introduce instruction-following artifacts that affect calibration metrics.

The analysis focus on three prompting strategies (zero-shot, role-based, chain-of-thought) without exploring tool use or alternative decoding methods.

\FloatBarrier
\section{Conclusion}

Language models now systematically outperform Brazilian students on standardized examinations, marking a threshold moment for educational technology in the Global South. The 270,900 model responses generated through Alvorada-Bench reveal both the promise and limits of current systems. Models achieve near-perfect accuracy on culturally specific humanities content, comprehending Machado de Assis and Brazilian constitutional history as readily as Shakespeare and American civics. This cultural fluency, emerging without targeted training, suggests that large-scale pre-training naturally captures diverse knowledge bases when sufficient Portuguese text is included.

Alvorada-Bench establishes that language models have crossed the threshold of educational competence in Brazilian Portuguese. The question is no longer whether these systems can handle Portuguese educational content, but how to deploy them equitably and effectively.

\bibliographystyle{unsrtnat}

\begin{thebibliography}{6}

\bibitem[Almeida et al.(2023)]{almeida2023bluexbenchmarkbasedbrazilian}
Thales Sales Almeida, Thiago Laitz, Giovana K. Bonás, and Rodrigo Nogueira.
\newblock BLUEX: A benchmark based on Brazilian Leading Universities Entrance eXams.
\newblock \emph{arXiv preprint arXiv:2307.05410}, 2023.
\newblock URL \url{https://arxiv.org/abs/2307.05410}.

\bibitem[OpenAI(2024)]{openai2024gpt4technicalreport}
OpenAI.
\newblock GPT-4 Technical Report.
\newblock \emph{arXiv preprint arXiv:2303.08774}, 2024.
\newblock URL \url{https://arxiv.org/abs/2303.08774}.

\bibitem[Gupta et al.(2025)]{gupta2025multilingualperformancebiaseslarge}
Vansh Gupta, Sankalan Pal Chowdhury, Vilém Zouhar, Donya Rooein, and Mrinmaya Sachan.
\newblock Multilingual Performance Biases of Large Language Models in Education.
\newblock \emph{arXiv preprint arXiv:2504.17720}, 2025.
\newblock URL \url{https://arxiv.org/abs/2504.17720}.

\bibitem[Guo et al.(2025)]{guo2025largelanguagemodelsenglish}
Yanzhu Guo, Simone Conia, Zelin Zhou, Min Li, Saloni Potdar, and Henry Xiao.
\newblock Do Large Language Models Have an English Accent? Evaluating and Improving the Naturalness of Multilingual LLMs.
\newblock \emph{arXiv preprint arXiv:2410.15956}, 2025.
\newblock URL \url{https://arxiv.org/abs/2410.15956}.

\bibitem[Hou and Ji(2024)]{hou2024comparing}
Wenpin Hou and Zhicheng Ji.
\newblock Comparing Large Language Models and Human Programmers for Generating Programming Code.
\newblock \emph{Advanced Science}, 12(8), 2024.
\newblock DOI: \url{10.1002/advs.202412279}.

\bibitem[Locatelli et al.(2024)]{locatelli2024examiningbehaviorllmarchitectures}
Marcelo Sartori Locatelli, Matheus Prado Miranda, Igor Joaquim da Silva Costa, Matheus Torres Prates, Victor Thomé, Mateus Zaparoli Monteiro, Tomas Lacerda, Adriana Pagano, Eduardo Rios Neto, Wagner Meira Jr., and Virgilio Almeida.
\newblock Examining the Behavior of LLM Architectures Within the Framework of Standardized National Exams in Brazil.
\newblock \emph{arXiv preprint arXiv:2408.05035}, 2024.
\newblock URL \url{https://arxiv.org/abs/2408.05035}.

\end{thebibliography}

\appendix

\section{Prompt Templates}

\subsection{Zero-Shot Prompt Template}

\begin{promptbox}[Zero-Shot Prompt]
**REGRA: Responda SOMENTE com JSON valido. Nenhum outro texto e permitido.**

Responda a seguinte questao de multipla escolha selecionando uma opcao.

Pergunta: {question}

Alternativas:
{options}

Formato JSON exigido:
{"resposta": "letra", "dificuldade": 0-10, "confianca": 0-10, "bloom": "Remember/Understand/Apply/Analyze/Evaluate/Create"}

**IMPORTANTE: Toda a sua resposta deve ser somente JSON valido.** 
\end{promptbox}

\subsection{Role-Playing Prompt Template}

This prompt establishes a specific persona of a high-performing Brazilian student, providing exam-specific context to potentially activate relevant knowledge patterns.

\begin{promptbox}[Role-Playing Prompt]
**REGRA: Responda SOMENTE com JSON valido. Nenhum outro texto e permitido.**

Voce e um estudante brasileiro de alto rendimento que se destacou em vestibulares como ENEM, FUVEST, ITA e UNICAMP. Voce possui amplo conhecimento em todas as disciplinas e habilidades excepcionais de realizacao de provas. 

Como esse estudante especialista, responda a seguinte questao de multipla escolha com a confianca e a metodologia que o tornaram bem-sucedido.

Pergunta: {question}

Alternativas:
{options}

Aplique sua estrategia especialista de resolucao de provas:
- Use seu profundo conhecimento da disciplina
- Aplique as tecnicas de eliminacao que voce dominou
- Considere padroes e pegadinhas tipicos dos exames
- Baseie-se em sua experiencia com questoes semelhantes
- Avalie a dificuldade da questao e seus requisitos cognitivos

Formato JSON exigido:
{"resposta": "letra", "dificuldade": 0-10, "confianca": 0-10, "bloom": "Remember/Understand/Apply/Analyze/Evaluate/Create"}

**IMPORTANTE: Toda a sua resposta deve ser somente JSON valido.** 
\end{promptbox}

\subsection{Chain-of-Thought Prompt Template}

\begin{promptbox}[Chain-of-Thought Prompt]
**REGRA: Responda SOMENTE com JSON valido. Nenhum outro texto e permitido.**

Responda a seguinte questao de multipla escolha usando raciocinio passo a passo.

Pergunta: {question}

Alternativas:
{options}

Pense nisso de forma sistematica:
1. Primeiro, identifique o que a questao esta pedindo
2. Decomponha os conceitos-chave ou as informacoes fornecidas
3. Analise cada alternativa em relacao aos requisitos da questao
4. Elimine alternativas incorretas com justificativa
5. Selecione a melhor resposta e avalie suas caracteristicas

Formato JSON exigido:
{"resposta": "letra", "dificuldade": 0-10, "confianca": 0-10, "bloom": "Remember/Understand/Apply/Analyze/Evaluate/Create"}

**IMPORTANTE: Toda a sua resposta deve ser somente JSON valido.** 
\end{promptbox}

\section{Bloom's Taxonomy Mapping}

Models were instructed to classify questions according to the revised Bloom's taxonomy:
\begin{itemize}
\item \textbf{Remember}: Recall facts and basic concepts
\item \textbf{Understand}: Explain ideas or concepts
\item \textbf{Apply}: Use information in new situations
\item \textbf{Analyze}: Draw connections among ideas
\item \textbf{Evaluate}: Justify a stand or decision
\item \textbf{Create}: Produce new or original work
\end{itemize}

\section{Model Specifications}

\subsection{Complete Model Specifications}

Twenty language models were evaluated in August 2025. The table below provides complete specifications for reproducibility.

\begin{table}[htbp]
\centering
\begin{tabular}{llll}
\toprule
Provider & Model & Context Length \\
\midrule
Anthropic & Claude 3.5 Haiku & 200K \\
Anthropic & Claude 3.5 Sonnet & 200K \\
Anthropic & Claude 3.7 Sonnet & 200K \\
Anthropic & Claude 3 Opus20240229 & 200K \\
Anthropic & Claude 4 Opus & 1M \\
Anthropic & Claude 4 Sonnet & 1M \\
DeepSeek & DeepSeek Chat & 64K \\
DeepSeek & DeepSeek Reasoner & 64K \\
OpenAI & GPT-4.1 & 1M \\
OpenAI & GPT-4.1 mini & 1M \\
OpenAI & GPT-4.1 nano & 1M \\
OpenAI & GPT-4o & 128K \\
OpenAI & GPT-4o mini & 128K \\
OpenAI & o1 & 128K \\
OpenAI & o1 mini & 128K \\
OpenAI & o1 preview & 128K \\
OpenAI & o3 & 200K \\
OpenAI & o3 mini & 200K \\
OpenAI & o3 pro & 200K \\
OpenAI & o4 mini & 200K \\
\bottomrule
\end{tabular}
\caption{Complete Model Specifications}
\label{tab:model_specs}
\end{table}

\textbf{Notes:}
\begin{itemize}
\item All models were accessed via their respective official APIs
\item Context length indicates the maximum token window for input processing
\end{itemize}

\end{document}